\documentclass[journal]{IEEEtran}
\usepackage[noadjust]{cite}
\usepackage{enumitem}
\usepackage{amsmath, amssymb,bm}
\usepackage{booktabs}
\usepackage{graphicx,epstopdf,subfigure}
\usepackage{multirow}
\usepackage{tabularx,threeparttable}
\usepackage{array}
\usepackage{xcolor}
\usepackage{comment}
\usepackage[hidelinks]{hyperref}
 \usepackage[normalem]{ulem} 

\ifCLASSINFOpdf
\else
\fi
\usepackage{fancyhdr}

\begin{document}

\title{PnP-AdaNet: Plug-and-Play Adversarial Domain Adaptation Network with a Benchmark at Cross-modality Cardiac Segmentation}

%\title{Unsupervised Domain Adaptation of Cross-modality Heart Segmentation via Adversarial Learning}

\author{Qi Dou*, Cheng Ouyang*, Cheng Chen, Hao Chen, Ben Glocker, Xiahai Zhuang, and Pheng-Ann Heng

\thanks{The * indicates equal contribution.}
\thanks{Q. Dou* was with Department of Computer Science and Engineering, The Chinese University of Hong Kong, and now is with Department of Computing, Imperial College London.}
\thanks{C. Ouyang* was with Department of Electrical Engineering and Computer Science, University of Michigan, and now is with Department of Computing, Imperial College London.}
\thanks{C. Chen, H. Chen and P. A. Heng are with Department of Computer Science and Engineering, The Chinese University of Hong Kong.}
\thanks{H. Chen is also with Imsight Medical Technology, Co, Ltd., China.}
\thanks{B. Glocker is with Department of Computing, Imperial College London.}
\thanks{X. Zhuang is with School of Data Science, Fudan University, China.}
\thanks{Emails at: Q. Dou \textit{dqcarren@gmail.com} and X. Zhuang \textit{zxh@fudan.edu.cn}}
}

\markboth{}
{Shell \MakeLowercase{\textit{et al.}}: Bare Demo of IEEEtran.cls for Journals}

\maketitle

\begin{comment}
\thispagestyle{fancy}
\fancyhead{} 
\lhead{} 
\lfoot{} 
\cfoot{\small{Copyright \copyright~2016 IEEE. Personal use of this material is permitted. However, permission to use this material for any other \\  purposes must be obtained from the IEEE by sending a request to pubs-permissions@ieee.org.}} 
\rfoot{}
\end{comment}

\begin{abstract}
Deep convolutional networks have demonstrated the state-of-the-art performance on various medical image computing tasks.
Leveraging images from different modalities for the same analysis task holds clinical benefits.
However, the generalization capability of deep models on test data with different distributions remain as a major challenge.
In this paper, we propose the \textit{PnP-AdaNet} (plug-and-play adversarial domain adaptation network) for adapting segmentation networks between different modalities of medical images, e.g., MRI and CT.
We propose to tackle the significant domain shift by aligning the feature spaces of source and target domains in an unsupervised manner.
Specifically, a domain adaptation module flexibly replaces the early encoder layers of the source network, and the higher layers are shared between domains.
With adversarial learning, we build two discriminators whose inputs are respectively multi-level features and predicted segmentation masks.
We have validated our domain adaptation method on cardiac structure segmentation in unpaired MRI and CT.
The experimental results with comprehensive ablation studies demonstrate the excellent efficacy of our proposed \textit{PnP-AdaNet}.
Moreover, we introduce a novel benchmark on the cardiac dataset for the task of unsupervised cross-modality domain adaptation. We will make our code and database publicly available, aiming to promote future studies on this challenging yet important research topic in medical imaging.

\end{abstract}

\begin{IEEEkeywords}
Domain adaptation, adversarial network, cross-modality images, cardiac segmentation, benchmark.
\end{IEEEkeywords}

\IEEEpeerreviewmaketitle

\section{Introduction}
\label{sec:intro}

\IEEEPARstart{D}{eep} learning models, especially convolutional neural networks (CNNs), have achieved remarkable successes during the past years,
achieving state-of-the-art or even human-level performance on a variety of challenging medical imaging problems~\cite{esteva2017dermatologist,bejnordi2017diagnostic,de2018clinically}.
Typically, the deep networks are trained and tested on datasets where all the images are sampled from the same distribution.
Despite the risk of over-fitting, the models are able to produce highly-accurate predictions on new test data from the same domain.
However, it has been frequently observed that established models under-perform when being tested on samples coming from a related but different new target domain~\cite{torralba2011unbiased,kalogeiton2016analysing,tommasi2016learning}.
For medical image computing, the scenarios include that the test and training images come from different sites~\cite{gibson2018inter,bentaieb2018adversarial} or different scanning protocols~\cite{kamnitsas2017unsupervised,karani2018lifelong} or even different imaging modalities~\cite{dou2018unsupervised,jiang2018tumor}.

\begin{figure}
	\centering
	\includegraphics[width=0.5\textwidth]{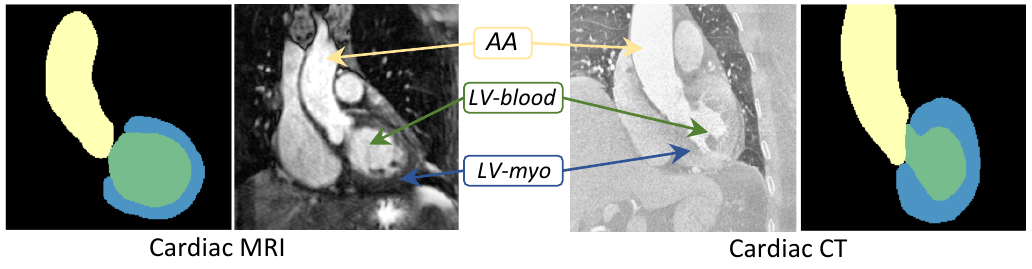}
	\vspace{-6mm}
	\caption{Illustration of severe domain shift existing in cross-modality images. The appearances of the cardiac structures (AA: ascending aorta, LV-blood: left ventricle blood cavity, LV-myo: left ventricle myocardium) look significantly different on MRI and CT images, though their segmentation masks are similar.}
	\vspace{-4mm}
	\label{fig:domain_shift}
\end{figure}

%In a broader view, the existence of domain shift is very common in real-world applications.
%We may easily encounter situations where the label space is identical (i.e., the analysis task is the same), while the data distributions are different.
Different from natural images which are generally obtained by optical cameras, a typical situation in the medical field is the usage of various imaging modalities, capturing different physical properties.
These different modalities play complementary roles in clinical procedure of disease diagnosis and treatment.
%In clinical practice, the diagnosis or treatment of diseases usually rely on multiple sources of images, which are complementary to each other.
%\textcolor{blue}{say something about heart disease diagnosis using both CT and MR}
For example, Magnetic Resonance Imaging (MRI) and Computed Tomography (CT) have become indispensable tools for cardiac imaging.
Specifically, MRI is ionizing radiation free and captures great contrast between soft tissues with high resolution in temporal space~\cite{mribook}.
It therefore can feature multi-parametric assessment of the myocardial contractivity and viability.
By contrast, CT allows rapid imaging of the cardiac morphology, myocardial viability and coronary calcification, and is with great spatial resolution~\cite{chartrand2007coronary}.
% but without requesting mental implantation free from the patients
%The MRI can feature free-ionizing multi-parametric assessment of the myocardial function and viability. The CT allows a rapid cardiac structure, morphology imaging with high spatial resolutions, in together with calcium scoring.
%It can be sometimes required to conduct the same image computation, 
In practice, often the same image analysis task is required, such as segmentation or quantification of cardiac structures, for both MRI and CT.
Considering that annotation is prohibitively time-consuming and expensive (e.g., a whole heart segmentation from either MRI or CT takes up to 8 hours by a well-trained operator~\cite{zhuang2013challenges}),
%In this case, considering that annotation is prohibitively time-consuming and expensive, 
effectively adapting the model trained on one modality to the other holds clinical benefits.
However, the appearances of cardiac MRI and CT are considerably different, with distinct contrast and intensity histogram, as shown in Fig.~\ref{fig:domain_shift}.
Unsupervised domain adaptation under such significant domain shift is very challenging, as well as yet to be explored.

%\textcolor{blue}{\uline{(xiahai: pls help refine above paragraph, especially blue parts, highlight why and how together use CT\&MRI in cardiology. Also pls add several references.)}}

Early studies on unsupervised domain adaptation focused on aligning the distributions in feature space, by minimizing measures of distances between the features extracted from the source and target domain.
For example, the Maximum Mean Discrepancy (MMD) was minimized together with a task-specific loss to learn domain-invariant and semantically-meaningful features in~\cite{tzeng2014deep}.
Long et al.~\cite{long2015learning} minimized MMD of domain features embedded in a reproducing kernel Hilbert space.
Sun et al.~\cite{sun2016deep} proposed to align the feature covariances between domains.
% the approach of~\cite{wang2017deep} minimized the domain differences furthermore based on both the first and second order information.
More recently, with the introduction of generative adversarial network (GAN) \cite{goodfellow2014generative} and its powerful extensions \cite{arjovsky2017wasserstein,CycleGAN2017}, the latent feature spaces across domains can be implicitly aligned via adversarial learning.
Notably, the DANN method was proposed to extract domain-invariant features by fully sharing weights of the CNN encoder between domains~\cite{ganin2016domain}.
Tzeng et al.~\cite{tzeng2017adversarial} introduced a more untied adversarial learning framework, named ADDA, where each domain has a dedicated encoder before the last classification layer.
Alternatively, in terms of GAN based domain adaptation, another stream of solution is to make use of image-to-image translation, i.e., training or testing the network with synthetic data, based on a CycleGAN~\cite{CycleGAN2017} foundation.

For medical image computing, adversarial learning has presented inspiring efficacy on a wide variety of tasks~\cite{pan2018synthesizing,fan2018adversarial,zhang2018translating,wolterink2017generative,zhang2018multi}.
In particular towards domain adaptation, promising studies have been increasingly conducted, with the aim of generalizing the learned models to unseen target domains.
For example, Zhang et al.~\cite{zhang2018task} transformed X-Ray images to appear like those source synthetic radiographs and directly tested the transformed images with the learned source model. 
Similarly based on the CycleGAN, Jiang et al.~\cite{jiang2018tumor} proposed a two-stage approach to first transform CT images to resemble MRI data, then conducted semi-supervised tumor segmentation with both synthetic data and a limited number of real MRI.
Meanwhile, following the spirit of aligning latent feature spaces, there are a set of works to extract domain-invariant representations.
Degel et al.~\cite{degel2018domain} minimized the segmentation loss with a domain discriminator to encourage feature domain-invariance across ultrasound datasets for left atrium segmentation. 
Ren et al.~\cite{ren2018adversarial} utilized adversarial learning to map the feature distribution of target images to the source domain for classifying histology images obtained in different staining procedures. 
%Dong et al.~\cite{dong2018unsupervised} discriminated segmentation predictions of the heart on both source and target X-Rays from those ground truth masks, based on the assumption that segmentation masks should be domain independent.
These works have demonstrated that imposing alignment in feature space
%either by inputting the feature activations or the segmentation masks to a domain discriminator, 
can help to generalize deep models to data from multiple sites.
One of the most related work is Kamnitsas et al.~\cite{kamnitsas2017unsupervised}, which conducted unsupervised domain adaptation by adversarial learning in multi-level feature space for brain lesion segmentation.
%The network learned domain-invariant features with an adversarial loss.
% serving as the supervision for representation extraction.
The experimental setting was challenging as an unseen MRI sequence was used in the target data.
By sharing encoders and aligning multi-level features, the method achieved promising results in target domain.
For a even more challenging setting of MRI and CT cross-modality image segmentation, to the best of our knowledge, related literature is limited.
Valindria et al.~\cite{valindria2018multi} developed a joint learning method for multi-organ segmentation using unpaired MRI and CT.
Zhang et al.~\cite{zhang2018translating} proposed cross-modality image translation for improving cardiac segmentation with synthetic data.
However, these works did not aim at the topic of unsupervised domain adaptation of CNNs at cross-modality medical images.

In this paper, we study the challenging issue of unsupervised cross-modality domain adaptation on multi-class segmentation problem.
%The significant shift between distributions of MRI and CT is to be tackled by our proposed unsupervised learning method.
We present a plug-and-play adversarial domain adaptation network, named \textit{PnP-AdaNet}, which effectively aligns the feature space of the target domain to that of the source domain. Specifically, the early encoders are replaced for target domain input, and higher layers are shared between domains.
At adversarial learning, we build two domain discriminators, respectively connecting multi-level features and predicted segmentation masks.
This paper is a substantial extension of our prior work~\cite{dou2018unsupervised}.
The main contributions in this paper are:

\begin{itemize}
	
	\item We tackle the task of unsupervised cross-modality domain adaptation for medical image segmentation. A novel \textit{PnP-AdaNet} is proposed to enable a flexible adaptation of the segmentation CNNs by plug-and-play feature encoders.
	
	%\item We pioneer the unsupervised cross-modality domain adaptation research for medical image segmentation with CNNs. A novel \textit{PnP-AdaNet} is proposed to enable flexible adaptation of the segmentation network by plug-and-play low-level feature encoders.
	
	\item We learn our model with unpaired MRI and CT images via adversarial learning. To enhance the supervision from discriminators, we aggregate multi-level features as well as segmentation mask predictions during training process.
	
	\item We extensively validate our method on multi-class cardiac segmentation. The mean Dice of four structures has been recovered from $13.2\%$ to $63.9\%$. We also conduct comprehensive ablation studies on key method components.
	
	\item To facilitate future studies on MRI and CT cross-modality domain adaptation, we introduce a new benchmark on our cardiac segmentation task, presenting performance of popular domain adaptation methods. 
	% We will make our code and database publicly available for ease of future investigations and comparisons.

	%We will release our code and pre-processed data for ease of direct comparisons.
	%Meanwhile, the original multi-modality cardiac dataset is available at the MICCAI 2017 MM-WHS challenge.
	
\end{itemize}

\section{Methods}

\begin{figure*}[t]
	\label{fig:overview}
	\centering
	\includegraphics[width=\textwidth]{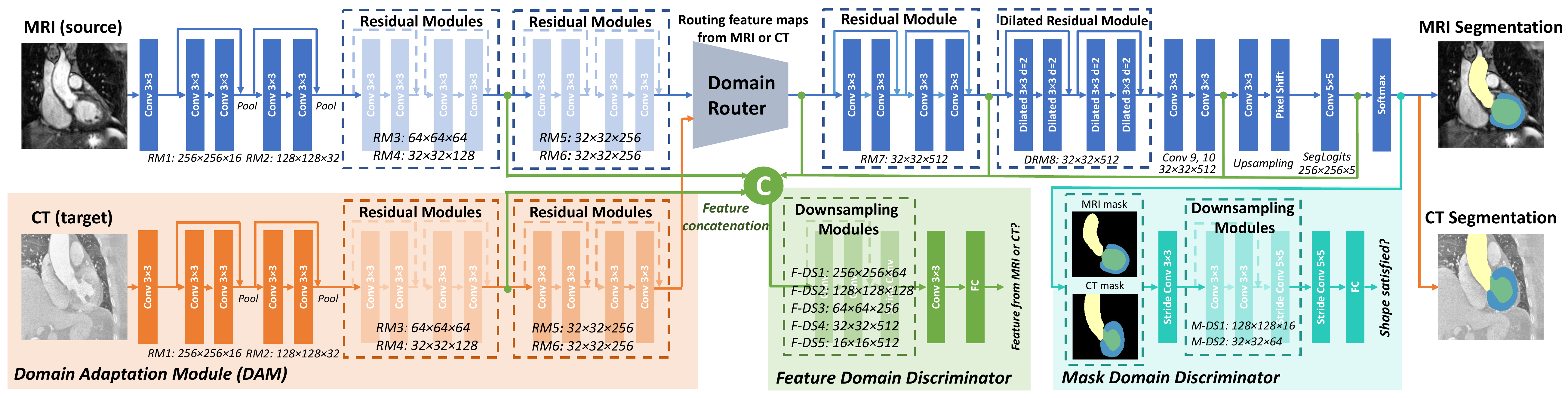}
	\vspace{-5mm}
	\caption{Overview of our proposed \textit{PnP-AdaNet} (plug-and-play adversarial domain adaptation network), consisting of a source segmentation network, a domain adaptation module (DAM) and two discriminators. Multi-level activations and predicted segmentation masks are aggregated for alignment of the latent feature space. The domain router is for testing. It chooses which set of early layers to connect to higher layers for segmentation task. Specifically, when testing source data, it chooses to use original source early layers; when testing target data, it chooses to use DAM layers.
	}
	%\vspace{-1mm}
\end{figure*}

The Fig.~2 is overview of our proposed \textit{PnP-AdaNet} method. 
With a standard segmentation CNN learned on source domain, we replace its early layers with a domain adaptation module while retain its higher layers, for testing on target domain data. Hence, we call our method as plug-and-play domain adaptation framework.
The adaptation module maps the target image to the distribution of source domain in latent feature space. This process is trained with adversarial loss in an unsupervised way. 

%Details of network architecture, adaptation mechanism, adversarial loss and training strategies are elaborated in the following subsections.

\subsection{Segmentation Network without Skip Connection}

The essence of our proposed \textit{PnP-AdaNet} is to establish an independent encoder for each domain and align their feature distributions in the latent space.
Considering that only the early layer encoders are updated while those higher layers are fixed, the feature spaces at different layers need to be self-contained, i.e., not mixed-up with each other.
This means that the network architectures using skip connections, e.g., the U-Net~\cite{ronneberger2015u} and DenseNet~\cite{yu2017automatic}, are not suitable choices.
Otherwise, the plug-and-play setting would be problematic, because those domain-specific low-level features would affect the aligned high-level feature space (which is supposed to be shared across domains).

In this regard, we set up our segmentation model as a dilated network~\cite{yu2017dilated}, which can extract representative features from a large receptive field, while also preserving the spatial acuity of feature maps.
Residual connections inside local scope are used for ease of gradients flow.
Specifically, as illustrated in Fig.~2,
the input image is firstly forwarded into a convolutional layer, then proceeded to a series of residual modules (termed as RM1-7, each consisting of the stacked $3 \! \times \! 3$ convolutions) and size down-sampled by a factor of eight.
%Next, another three RMs and one dilated RM are stacked to form a deep network.
Next, to enlarge the receptive field for extracting the contextual information, 4 dilated convolutional layers are used in DRM8 with a dilation factor of 2.
After another two convolutional layers Conv9 and Conv10, we conduct upsamling to get dense predictions for the segmentation task.
A $5 \! \times \! 5$ convolution operation is immediately followed to smooth out the activation maps.
Finally, a softmax layer is utilized to obtain probabilistic pixel-wise predictions.

%Formally, we denote the labeled dataset of $N^s$ samples from source domain by $X^s \! = \! \{(x_1^s,y_1^s),...,(x_{N^s}^s,y_{N^s}^s)\}$.
Formally, we denote the annotated dataset of the source domain by $X^s \! = \! \{(x_1^s,y_1^s),...,(x_i^s,y_i^s),...,(x_{N^s}^s,y_{N^s}^s)\}$,
where $x_i^s$ represents a sample pixel of the image and $y_i^s$ is its category label of anatomical structure.
We conduct supervised learning to establish a mapping $M^s$ from the input image to the label space.
The segmentation network of source domain is optimized by minimizing the hybrid loss $\mathcal{L}_\text{seg}$ composed of multi-class cross-entropy loss and Dice coefficient loss~\cite{milletari2016v}.
Moreover, we denote $y_{i,c}^s$ as binary label regarding class $c \! \in \! C$ for sample $x_i^s$, its probability prediction is $\hat{p}_{i,c}^s$, and the label prediction is $\hat{y}_{i,c}^s$, the overall segmentation loss function is:
%\vspace{-2mm}
\begin{equation}
%\small
\begin{split}
\mathcal{L}_{\text{seg}} % (Y^s, \hat{Y}^s)
= & - \sum \limits_{c \in C} \frac{ \sum_{i=1}^{N^s} 2 y_{i,c}^s \hat{y}_{i,c}^s }{  \sum_{i=1}^{N^s} y_{i,c}^s y_{i,c}^s + \sum_{i=1}^{N^s} \hat{y}_{i,c}^s \hat{y}_{i,c}^s } \\
& -\lambda \sum \limits_{i=1}^{N^s} \sum \limits_{c \in C} w^s_{c} \cdot y_{i,c}^s \log(\hat{p}_{i,c}^s) + \beta ||W||_2^2.
\end{split}
\end{equation}
The first term is a Dice loss for multiple anatomical structures, and the second term is cross-entropy loss for individual pixels.
The $w_{c}^s$ is the weighting factor which copes with the issue of class imbalance.
We combine the two complementary losses to tackle the challenging cardiac segmentation task.
In practice, we also tried just using either one, but the performance was not as high as using both.
The third term is the L2-regularization of the segmenter weights $W$.
The $\lambda$ and $\beta$ are trade-off weights.

For the ease of notation, we will omit the subscript index $i$ in following subsections, and directly use $x^s$ and $y^s$ to represent the sample and label from the source domain.

\subsection{Plug-and-Play Adaptation Mechanism}

After obtaining the segmentation network trained on the source domain, we next aim to adapt it onto the target domain, in an unsupervised manner.
%A straight-forward way is to fine-tune the model using labeled data from the new domain.
%However, this is obviously not the most favorable solution, because manual annotations are time-consuming as well as expensive.
%Investigating an unsupervised adaptation solution deserves devoted exploration efforts.
In conventional transfer learning, it is common to update the last several layers of the pre-trained network towards a new given task with a new label space.
The supporting assumption is that the early layers in the network extract low-level features which are universal for vision tasks.
The higher layers are more task-specific and learn semantic-level features for conducting the defined predictions~\cite{zeiler2014visualizing,yosinski2014transferable}.
In contrast, for domain adaptation, the defined task remains unchanged across domains.
This means that the label space for source and target domains are identical, e.g., we segment the same anatomical structures from MRI/CT images.
Basically, the distribution shift between the cross-modality domains are primarily low-level characteristics (e.g., gray-scale intensities) rather than high-level (e.g., geometric or semantic structures).
%Recent adversarial learning based domain adaptation methods were also developed on the basis of this fundamental assumption~\cite{ganin2016domain,tzeng2017adversarial}.

In these regards, for our model, we design a plug-and-play adaptation mechanism, i.e., a set of early layers being replaced, while the higher layers are reused for a new target domain. 
The underlying intuition is that, the higher layers are closely correlated with the shared semantic labels, while the respective early-layer encoders perform distribution mappings in feature space for our unsupervised domain adaptation.
Formally, the obtained source segmentation model $M^s$ is regarded as a layer-wise feature extractor composing stacked transformations of $\{M^s_{l_1},...,M^s_{l_n}\}$, with the subscript denoting the network layer index. The prediction of the semantic label is represented by:
%\vspace{1mm}
\begin{equation}
\hat{y}^s = M^s(x^s)= M^s_{l_1:l_n}(x^s) = M^s_{l_n}  \circ ... \circ M^s_{l_1} (x^s).
\vspace{1mm}
\end{equation}

For input $x^t  \! \! \in \! \! X^t$ from a target domain, we propose a domain adaptation module (DAM) denoted by $\mathcal{M}$ that maps $x^t$ to the feature space aligned with the source domain.
We denote the adaptation depth by $d$, i.e., the layers earlier than and including $l_d$ are replaced by DAM when processing the target domain images.
At the same time, the source model's upper layers are frozen during domain adaptation learning and reused for target inference.
Hence, the prediction for the target input is:
\begin{equation}
\vspace{1mm}
\begin{aligned}
\hat{y}^t = M^s_{l_{d+1}:l_n} \circ \mathcal{M}(x^t)= M^s_{l_n}  \circ ... \circ M^s_{l_{d+1}} \circ \mathcal{M} (x^t),
\end{aligned}
%\vspace{1mm}
\end{equation}
where $\mathcal{M}(x^t) = \mathcal{M}_{l_1:l_d}(x^t)  = \mathcal{M}_{l_d}  \circ ... \circ \mathcal{M}_{l_1} (x^t)$ represents the DAM which is a stack of convolutional layers as the feature encoder.
In practice, we set the DAM layer configurations the same as the replaced set of early layers of source model, i.e., $\{M^s_{l_1},...,M^s_{l_d}\}$.
This is a reasonable and safe implementation choice, as we can initialize the DAM with a pre-trained source encoder rather than random initializations.
This contributes to a more stable optimization at adversarial training, especially in our unsupervised learning scenario. 

Overall, we can find that the proposed plug-and-play domain adaptation mechanism is elegant and rather flexible at testing.
During inference for the target domain, the DAM directly replaces the early $d$ layers of the source domain network.
The images of the target domain are processed and mapped to the feature space of source domain using this DAM.
These adapted features are robust to the cross-modality domain shift, and can be correctly transformed into the label space via the fixed high-level layers.
It is also necessary to mention that our plug-and-play domain adaptation procedure does not hurt the performance for the source domain.
The early layer encoding path for the source domain is preserved and the higher layers are unaffected at learning.
Therefore, our \textit{PnP-AdaNet} can be flexibly tested for both target and source domain data, just by selecting the input path.

\subsection{Adversarial Learning for Feature Space Alignment}

We train our plug-and-play domain adaptation network via adversarial learning in an unsupervised manner.
In the spirit of GAN, a generator and a discriminator form a minimax two-player game.
The generator aims to capture the distribution of the real data, while the discriminator should identify whether a presented sample comes from the real or learned distributions.
%In previous literatures, the inputs to the discriminators can be either high-dimensional or low-dimensional data.
%These two modules, i.e., the generator and the discriminator, are alternatively optimized and they compete with each other, until the generator is able to successfully produce real-like data that the discriminator fails to distinguish.
In our \textit{PnP-AdaNet}, the DAM serves as the generator which maps the input target image into the latent feature space of the source domain.
The aim of the domain adaptation module is to encode representations which are aligned with those encoded from the source domain images.
Hence, the fixed layers in the higher part of the source network can be reused to make semantic-level predictions of segmentation masks.
As we have zero annotation for the target domain, the adaptation process is implicitly supervised by the discriminators, i.e., forming an adversarial learning game.

In our framework, we propose to employ two discriminators.
Specifically, the input to the first discriminator (i.e., the green part in Fig.~2) is an array of aggregated feature maps of the segmenter.
This input has a high dimension with a relatively complicated distribution.
A natural thinking here is that, we connect the output features of the DAM into the discriminator.
However, the convolutional neural network has a hierarchical architecture, and the features at one certain layer rely on activations from its previous layers, and also, the features are proceeded to affect following layers.
If we just monitor the encoded features immediately obtained from the DAM, the latent space alignment can be unstable.
In other words, we have no idea whether those activations in the layers earlier to the adaptation depth are aligned.
Also, the small shift that may still exist at the adaptation layer $l_d$ would be magnified after composing them in higher layers.
To overcome this problem, we aggregate the activations from multiple layers as input to the discriminator.
The gradients from the discriminator can flow to the DAM via multiple paths, so that the feature space alignment can be more tightly supervised.
This way of learning shares the spirit of deep supervision~\cite{lee2015deeply} to some extent.

In practice, we aggregate the activations from multiple levels of layers,
%an early encoder layer, the adaptation-depth layer and frozen higher layers,
and reshape them to the same resolution for channel concatenation.
Formally, we refer to the feature maps in the selected frozen layers as the set of $F_H (\cdot)$ where $H \! = \! \{k,... ,q\}$ being the set of selected layer indices.
Similarly, we denote the selected feature maps of DAM by $\mathcal{M}_A( \cdot )$ with $A$ being the selected layer set.
In this way, the feature space of the target domain is $(\mathcal{M}_A(x^t), F_H(x^t))$ and the array $(M^s_A(x^s), F_H(x^s))$ is their counterpart for the source domain.
Given the distribution of $(M^s_A(x^s), F_H(x^s)) \! \sim \! \mathbb{P}_{\text{feature}}^s$ and that of $(\mathcal{M}_A(x^t), F_H(x^t)) \! \sim \! \mathbb{P}_{\text{feature}}^t$, the distance between these two domain distributions which is to be minimized is represented as $W(\mathbb{P}_{\text{feature}}^s,\mathbb{P}_{\text{feature}}^t)$.
For stabilized training, we use Wassertein distance~\cite{arjovsky2017wasserstein} of the two distributions as:
\begin{equation}
%\small
%\vspace{-1mm}
W(\mathbb{P}_{\text{feature}}^s, \mathbb{P}_{\text{feature}}^t)=\inf \limits_{\gamma \sim \prod(\mathbb{P}_{\text{feature}}^s, \mathbb{P}_{\text{feature}}^t)} \mathbb{E}_{(\text{x},\text{y})\sim\gamma}[\Vert \text{x} - \text{y} \Vert],
\label{equ:emdis}
%\vspace{-1mm}
\end{equation}
where $\prod(\mathbb{P}_{\text{feature}}^s, \mathbb{P}_{\text{feature}}^t)$ is the set of all joint distributions of $\gamma(\text{x},\text{y})$ whose marginals are respectively $\mathbb{P}_{\text{feature}}^s$ and $\mathbb{P}_{\text{feature}}^t$.
%For the first discriminator, the input feature maps are first resized to the same resolution for concatenation of channels.

Aligning the latent feature space by directly inputting the high-dimensional activations to a discriminator is effective and essential.
This might be fine for classification tasks, but may be sub-optimal for segmentation tasks which require pixel-wise predictions with fine structures. Early studies using the GANs for segmentation applications (not necessarily under the domain adaptation setting) commonly input the predicted segmentation masks to the discriminator.
When the shape or structure of the predicted segmentation mask looks distorted (i.e., not looking like the real mask), the discriminator would impose a penalty.
For the specific problem of domain adaptation at segmentation, we also consider that monitoring the shape of the predicted segmentation mask is important.

To this end, we further include an auxiliary discriminator in our \textit{PnP-AdaNet}, whose inputs are the predicted segmentation masks of the source and target domains.
In this case, the input is more compact and has clearer semantic meaning, compared with that of the first discriminator.
We denote the segmentation predictions for the target and source domains by $\mathcal{S}(x^s) \! \sim \! \mathbb{P}^s_{\text{mask}}$ and $\mathcal{S}(x^t) \! \sim \! \mathbb{P}^t_{\text{mask}}$. Following the Eq.~(\ref{equ:emdis}), we also employ the Wassertein distance between source and target distributions:
\begin{equation}
\small
\vspace{-1mm}
W(\mathbb{P}_{\text{mask}}^s, \mathbb{P}_{\text{mask}}^t)=\inf \limits_{\gamma \sim \prod(\mathbb{P}_{\text{mask}}^s, \mathbb{P}_{\text{mask}}^t)} \mathbb{E}_{(\text{x},\text{y})\sim\gamma}[\Vert \text{x} - \text{y} \Vert].
\label{equ:emdis_d2}
%\vspace{-1mm}
\end{equation}

%The two discriminators use the same network configuration in practice.
%Specifically, our constructed model consists of five residual blocks, one convolutional layer and a fully-connected layer, as illustrated in Fig.~2.
%In every block, the number of channels is doubled while the feature map sizes are decreased.
The detailed network architectures of the discriminators are illustrated in Fig.~2.
For the model configuration, the feature discriminator is relatively deeper than the mask discriminator.
%We employ strided convolutions for down-sampling and Leaky\_ReLU as non-linearity for stable training.

\subsection{Loss Functions and Training Strategies}
\label{subsec:training}

In adversarial learning, the DAM is pitted against an adversary with the above two discriminators.
We represent the first discriminator with $\mathcal{D}_f$ which takes high-dimensional features as input,
and denote the second discriminator by $\mathcal{D}_m$ which takes compact predicted segmentation masks as input. 
The pair of $\mathcal{D}_f$ and $\mathcal{D}_m$ implicitly estimate the $W(\mathbb{P}_{\text{feature}}^s,\mathbb{P}_{\text{feature}}^t)$ and the
$W(\mathbb{P}_{\text{mask}}^s,\mathbb{P}_{\text{mask}}^t)$, respectively. 
During the learning process, the discriminators would try to differentiate the inputs from the source and target domains.
The domain adaptation with DAM aims to not only remove the domain-specific patterns in early layers, but also disallow their recovery at higher semantic-level layers. The generator $\mathcal{M}$ (DAM) and the discriminators $\{\mathcal{D}_f$, $\mathcal{D}_s\}$ are jointly optimized via adversarial loss functions. Specifically, the loss for generator DAM is:
\begin{equation}
\small
\begin{split}
\mathcal{L}_{\mathcal{M}} \! = \! & -\mathbb{E}_{ (\mathcal{M}_{A}(x^t), F_{H}(x^t)) \sim \mathbb{P}_{\text{feature}}^t} [\mathcal{D}_f(\mathcal{M}_{A}(x^t), F_{H}(x^t))] \\
& -\mathbb{E}_{ \mathcal{S}(x^t) \sim \mathbb{P}_{\text{mask}}^t} [\mathcal{D}_m(\mathcal{S}(x^t))]. 
\label{equ:Gloss}
\end{split}
\end{equation}
%Moreover, with the $x^s \! \! \sim \! \! X^s$ representing the set of source data, 
The losses for learning the discriminators $\mathcal{D}_f$ and $\mathcal{D}_m$ are:
\begin{equation}
\small
\begin{split}
\mathcal{L}_{\mathcal{D}_f} = &~ \mathbb{E}_{(\mathcal{M}_{A}(x^t), F_{H}(x^t))  \sim \mathbb{P}_{\text{feature}}^t}[\mathcal{D}_f(\mathcal{M}_{A}(x^t), F_{H}(x^t))] \ - \\
&~\mathbb{E}_{ (M^s_{A}(x^s), F_{H}(x^s))\sim \mathbb{P}_{\text{feature}}^s}[\mathcal{D}_f(M^s_{A}(x^s),F_{H}(x^s))], \\
&~  s.t. \  \Vert \mathcal{D}_f \Vert_{L}\leq K,
\end{split}
\label{equ:Dloss_1}
\end{equation}

\begin{equation}
\small
\begin{split}
\mathcal{L}_{\mathcal{D}_m} = &~ \mathbb{E}_{\mathcal{S}(x^t)  \sim \mathbb{P}_{\text{mask}}^t}[\mathcal{D}_m(\mathcal{S}(x^t))]  - 
\mathbb{E}_{\mathcal{S}(x^s)  \sim \mathbb{P}_{\text{mask}}^s}[\mathcal{D}_m(\mathcal{S}(x^s))], \\
&~  s.t. \  \Vert \mathcal{D}_m \Vert_{L}\leq K,
\end{split}
\label{equ:Dloss_2}
\end{equation}
where $K$ is the Lipschitz constraint of $\mathcal{D}_f$, $\mathcal{D}_m$.
By alternatively updating the generators and discriminators,
% $\mathcal{M}$, $\mathcal{D}_f$ and $\mathcal{D}_m$, 
%the discriminators output more precise estimations of Wassertein distances between the distributions of source and target domains.
%Also, in the meanwhile, 
the DAM becomes more effective to generate source-like features from target data for domain adaptation.
%Overall, the loss function is as follows, with $\lambda_f$ and $\lambda_m$ being trade-off parameters:

%\begin{equation}
%\mathcal{L} = \mathcal{L}_{\mathcal{M}} + \lambda_{f} \mathcal{L}_{\mathcal{D}_f} + \lambda_{m} \mathcal{L}_{\mathcal{D}_m} + \gamma ||\theta_2||^2_2.
%\end{equation}

In practice, we first train the segmentation network on the source domain in a supervised manner with standard stochastic gradient descent.
We employ the Adam optimizer with a batch size of 10 and a learning rate of $1 \! \times \! 10^{-3}$.
After obtaining the segmenter, we train $\{\mathcal{M}, \mathcal{D}_f, \mathcal{D}_m\}$ with above adversarial loss for unsupervised domain adaptation.
To begin with, we solely update the discriminators for 20k iterations with a batch size of 6 as a pre-training process. Next, we alternatively update the generator and discriminators. Following the heuristic rules of training WGAN~\cite{arjovsky2017wasserstein}, we update the generator $\mathcal{M}$ for once, every 20 iterations updating the $\mathcal{D}_f$ and $\mathcal{D}_m$.
In adversarial learning, we use the RMSProp optimizer with a learning rate of $3 \! \times  \!10^{-4}$, and a stepped decay rate of $0.98$ every $100$ joint updates. Weight clipping for discriminator weights is $0.03$.
The discriminator and generator losses are scaled with a factor of 0.002.
Dropout ($\text{rate} \! = \!0.25$) and batch normalization are used in all the convolutional layers.

As learning with the adversarial loss is notoriously difficult, we carefully adjust the implementation settings to stabilize the optimization process.
Specifically, to avoid instability resulting from the sparse gradient, we use Leaky-ReLU as the activation function and strided convolutions for down-sampling inside the discriminators, instead of using the common practice of ReLU and max-pooling. 
% as further reduction of the learning rate~\cite{nagarajan2017gradient}. 

\section{Benchmark Dataset and Evaluation Metrics}
\label{sec:dataset}

\subsection{Benchmark Dataset for Cross-modality Domain Adaptation}

Publicly available medical datasets which contain different modalities of images for the same anatomical structure are rare, hindering progress of investigating the challenging task of MRI and CT cross-modality domain adaptation.
Fortunately, the challenge of \textit{MICCAI 2017 Multi-Modality Whole Heart Segmentation (MM-WHS)} presented 20 MRI and 20 CT cardiac images with accurate manual segmentation annotations.
%These multi-modal images are unpaired, with the MRI data coming from the Imaging Division at King's College London in UK, and the CT data coming from the Shang-hai Shuguang Hospital in China.
The images are unpaired with the MRI data and CT data coming from different patients and different sites.
We refer the readers to the original data description paper of Zhuang et al.~\cite{zhuang2016multi} for more details about data acquisition such as the employed scanning protocols.
For evaluating the domain adaptation on cardiac segmentation, we include following four structures: ascending aorta (AA), the left atrium blood cavity (LA-blood), the left ventricle blood cavity (LV-blood), and the myocardium of the left ventricle (LV-myo).
We randomly split each modality of the data into training (16 subjects) and testing (4 subjects) subsets in experiments.

For the setting of cross-modality domain adaptation of deep networks, we conduct pre-processings of the data.
Specifically, the MRI and CT images are reoriented, resized and cropped centering at the heart region, such that the view of multi-modal images are roughly on the same page.
%We crop the images into the size of $256 \! \times \! 256 $ at coronal plane, centering at the heart region. 
We extract MRI and CT scans by 2D slices of size $256\times256$ at coronal plane during training, and the models input three adjacent slices for contextual information.
The mask label of the middle slice is used as the ground truth.
We conduct the intensity normalization in 3D for each modality, respectively.
Augmentations of rotation, zooming and affine transformations are utilized to enlarge the training database.
We plan to make our pre-processed database publicly available, to facilitate future studies on cross-modality domain adaptation using this dataset.

\begin{comment}
Fort the setting of cross-modality domain adaptation of deep networks, we conduct pre-processings of the data.
Specifically, the CT images are resampled to a voxel spacing of $1 \!\times \! 1 \times \! 1 \text{mm}^3$.
The MRI images are kept as their original voxel spacing which is already around $1 \!\times \! 1 \times \! 1 ~\text{mm}^3$.
%Specifically, all the MRI and CT images are re-sampled to a voxel spacing in the range of $0.8$ to $1.2$~mm.
The MRI and CT images are re-oriented into the same orientation.
We crop the images into the size of $256 \! \times \! 256 $ at coronal plane, centering at the heart region. 
We extract the MRI and CT scans by 2D slices during the training, i.e., the model inputs three adjacent slices along the coronal plane.
The mask label of the middle slice is used as the ground truth.
We conduct the intensity normalization in 3D for each modality, respectively.
Augmentations of rotation, zooming and affine transformations are utilized to enlarge the training database.
We plan to make our pre-processed database publicly available, to facilitate future studies on cross-modality domain adaptation on this dataset.
\end{comment}

\subsection{Evaluation Metrics on Segmentation Performance}

For the evaluation metrics, we follow the common practice to quantitatively evaluate segmentation methods~\cite{dou20173d}.
The Dice coefficient $\!([\%])\!$ is used to assess the agreement between the predicted segmentation and ground truth for cardiac structures.
We use the symmetric average surface distance (ASD)$[\text{voxel}]$) to measure the segmentation performance from the perspective of boundary agreement.
A higher Dice and a lower ASD indicate a better segmentation performance.
Both metrics are presented in the format of \emph{mean$\pm$std}, which shows the average performance with the cross-subject variations of the results.
For some results, the N/A on the ASD means that at least one subject did not receive any correct prediction on the structure.

\section{Experiments}
\vspace{2mm}
\label{sec:exp}

\begin{figure*}[!htp]
	\label{fig:cmp}
	\centering
	\includegraphics[width=1.0\textwidth]{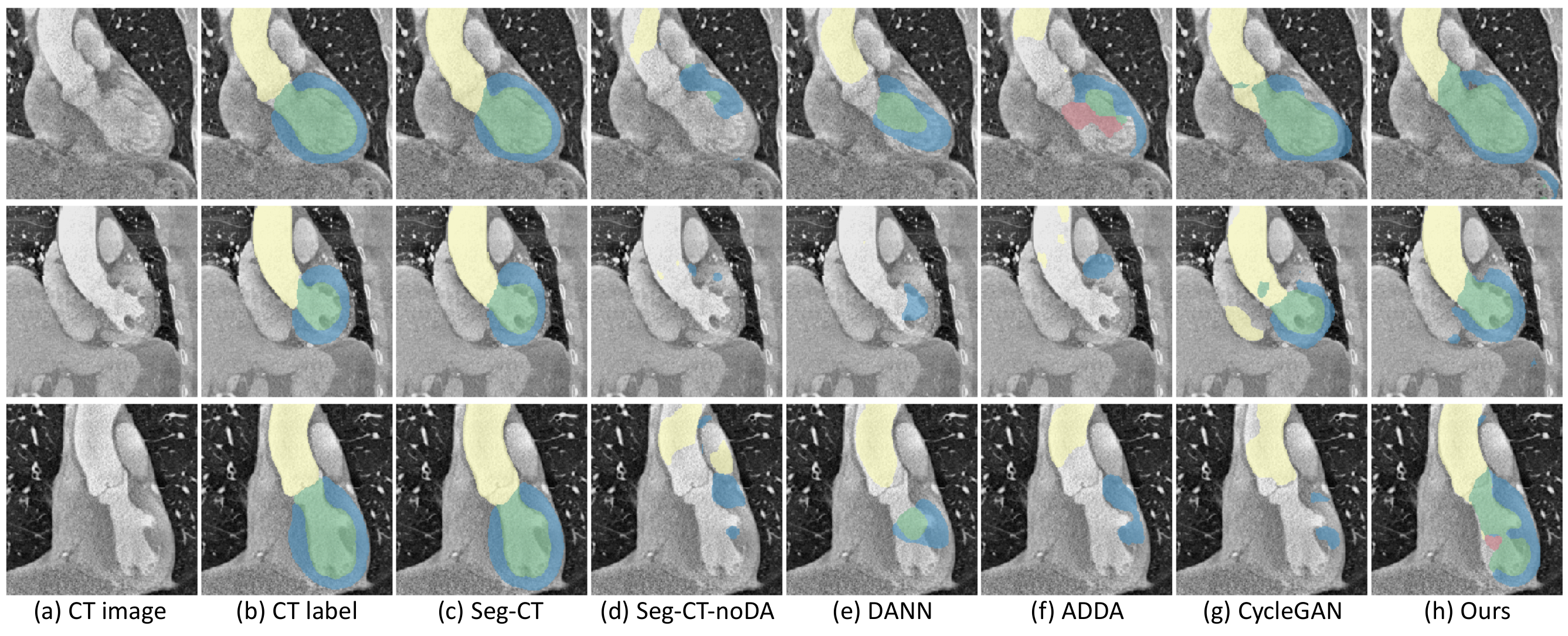}
	\vspace{-6mm}
	\caption{Results of different methods for CT image segmentations. Each row presents one typical example, from left to right: (a) raw CT images (b) ground truth labels (c) supervised training on CT (d) directly applying MRI segmenter on CT data (e) results of DANN (f) results of ADDA (g) results of CycleGAN (h) results of our proposed \textit{PnP-AdaNet}. The structures of AA, LA-blood, LV-blood and LV-myo are indicated by yellow, red, green and blue colors, respectively.}
	%\vspace{-2mm}
\end{figure*}

\subsection{Experimental Settings}
In our experiments, we first set the source domain as MRI and target domain as CT.
We conducted extensive experiments to demonstrate the severe cross-modality domain shift and the effectiveness of domain adaptation strategies.
Specifically, we designed the following experimental configurations:\\
1) training and testing the segmentation network on the source domain (referred as \textit{Seg-MRI});\\
2) training and testing the segmentation network on annotated target domain images, as an upper bound (referred as \textit{Seg-CT}); \\
%3) fine-tuning the source domain segmentation network using annotated target domain images, i.e., supervised transfer learning (referred as \textit{Seg-CT-STL}), and testing on target domain; \\
3) directly testing the source domain segmenter on target data, with no domain adaptation (referred as \textit{Seg-CT-noDA}); \\
4) our \textit{PnP-AdaNet} for unsupervised domain adaptation.

%In addition, we also referenced the performance of the state-of-the-art on the task of whole-heart segmentation.

In addition, we optimized the practical configurations of our \textit{PnP-AdaNet}. 
Specifically, our ablation studies investigated: i) how the balancing ratio between the two discriminators affects the domain adaptation performance,
and ii) the importance of inputing multiple levels of feature maps to $\mathcal{D}_f$.
Moreover, we implemented a series of existing popular unsupervised domain adaptation methods on our dataset, and provided a benchmark for the cross-modality
MRI to CT domain adaptation.
Last but not least, we also report the segmentation results by reverting the adaptation direction, i.e., adapting from CT to MRI.

\begin{table}[t]
	\label{tab:results1-small}
	\centering
	\scriptsize
	\renewcommand{\arraystretch}{1.2}
	\caption{Quantitative evaluation with Dice of segmentation results on cardiac structures of different experimental configurations.}
	%\vspace{-2mm}
	\begin{tabular} { |m{2.62cm}|m{1.0cm}|m{1.0cm}|m{1.0cm}|m{1.0cm}| }
		\hline
		Experiment configurations              & ~~~~AA          & LA-blood        & LV-blood      & ~~LV-myo        \\
		%\cline{2-9}
		%&~~~ Dice & ~~~ ASD & ~~~ Dice & ~~~ ASD & ~~~ Dice & ~~~ ASD & ~~~ Dice  & ~~~ ASD \\
		\hline
		
		Cascaded-FCN-MRI~\cite{payermulti}     & 76.6$\pm$13.8  & 81.1$\pm$13.8   & 87.7$\pm$7.7   & 75.2$\pm$12.1  \\
		U-Net-MRI~\cite{ronneberger2015u}      & 70.6$\pm$20.2  & 74.0$\pm$24.7   & 81.1$\pm$23.8  & 68.1$\pm$25.3 \\
		Cascaded-FCN-CT~\cite{payermulti}      & 91.1$\pm$18.4  & 92.4$\pm$3.6    & 92.4$\pm$3.3   & 87.2$\pm$3.9   \\
		U-Net-CT~\cite{ronneberger2015u}       & 94.0$\pm$6.2   & 91.0$\pm$5.2    & 91.0$\pm$4.3   & 86.1$\pm$4.2   \\
		\hline
		Seg-MRI                                & 80.3$\pm$1.8   & 78.1$\pm$8.0    & 88.3$\pm$2.1   & 70.8$\pm$2.1  \\
		Seg-CT                                 & 77.6$\pm$29.2  & 89.4$\pm$2.2    & 91.3$\pm$2.5   & 85.8$\pm$1.5   \\
		\hline
		Seg-CT-noDA                            & 31.5$\pm$23.9  & ~2.7$\pm$0.8    & ~3.4$\pm$5.8    & 15.3$\pm$17.2   \\
		\hline
		Proposed \textit{PnP-AdaNet}                    & 74.0$\pm$7.3   & 68.9$\pm$5.2    & 61.9$\pm$10.7  & 50.8$\pm$7.0   \\
		\hline
		%\multirow{2}{*}{Experimental Configurations} & \multicolumn{5}{c|}{Average Surface Distance (ASD) [voxel]} \\
		%\cline {2-6}
		%                                       & ~~~~~~AA          & ~~LA-blood        & ~~~LV-blood      & ~~~LV-myo       & ~~~ Mean  \\
		% \hline
		%Seg-MRI                                & ~~~6.1$\pm$2.5   & ~~~6.4$\pm$4.0    & ~~~2.8$\pm$1.2   & ~~~2.8$\pm$1.5  & ~~~4.5$\pm$2.3  \\
		%Seg-CT                                 & ~~~2.6$\pm$2.3   & ~~~3.3$\pm$1.3	  & ~~~1.7$\pm$0.2   & ~~~2.8$\pm$0.9  & ~~~2.6$\pm$1.2  \\
		%\hline
		%Seg-CT-noDA                            & ~~21.4$\pm$14.1  & ~19.6$\pm$4.6     & ~~~~~N/A         & ~~20.7$\pm$7.7  & ~~~~~~NA     \\
		%\hline
		%Proposed PaP-AdaNet                    & ~~12.8$\pm$3.2   & ~~6.3$\pm$2.3     & ~~17.4$\pm$7.0   & ~~14.7$\pm$4.8  & ~~12.8$\pm$4.3 \\
		
		%\hline
	\end{tabular}
\end{table}

\subsection{Results of PnP-AdaNet for Domain Adaptation}

%The results of different configurations are listed in Table~1, which presents the effectiveness of our PaP-AdaNet for unsupervised domain adaptation on cross-modality medical data.

Our employed segmentation network without skip connection serves as the basis for our subsequent domain adaptation procedures.
Hence, we choose to first validate the performance of this dilated network architecture on the applied whole-heart segmentation task.
Specifically, for \textit{Seg-MRI} setting, our model achieves an average Dice of $79.4\%$ across four structures.
%For the \textit{Seg-CT}, our segmentation network obtains an average Dice of $86.0\%$ with an ASD of $2.6$ mm.
To see the comparison, we reference Payer et al.~\cite{payermulti} which used two cascaded fully convolutional networks and achieved an average Dice of 80.2\%, ranking the first in \textit{MICCAI 2017 MM-WHS Challenge}.
We also directly quote their reported U-Net results as a complementary comparison.
%Only the Dice coefficients were reported in~\cite{payermulti}.
As the detailed results listed in Table~1, our network's segmentation performance with standard training is comparable to the state-of-the-arts, which employed networks based on 3D convolutional kernels. 
Hence, we can safely regard our trained segmentation network as a standard baseline model for the cardiac segmentation task.
%Then, we can do our main experiments based on this network, to explore the goal of cross-modality domain adaptation and validate the effectiveness of our proposed PaP-AdaNet.  

As for observing the severe domain shift inherent in cross-modality medical images, we first directly deploy the segmentation model trained on MRI domain to CT data.
%without any domain adaptation procedure. 
Unsurprisingly, the MRI segmenter completely fails on CT images, with an average Dice of merely 13.2\% across all the four structures. 
Specifically, the \textit{Seg-CT-noDA} only receives a Dice of 2.7\% for LA-blood and 3.4\% for LV-blood.
%The model did not even output any correct predictions for a subject on LV-blood.
%These observations convey that, although the cardiac MRI and CT images present the anatomical maps of the same organ, their significant differences in appearance would make it impossible for a MRI segmenter to extract effective features which also directly work on CT.
After domain adaptation, our \textit{PnP-AdaNet} presents a great recovery of the segmentation performance on target CT data compared with \textit{Seg-CT-noDA}. 
More specifically, our method has increased the average Dice across the four cardiac structures by 50.7\%, achieving a score of 63.9\%. 
As presented in Fig.~3, the predicted segmentation masks from \textit{PnP-AdaNet} can successfully localize the cardiac structures and further capture their anatomical shapes.
Notably, the segmentation performance on aorta has been significantly recovered after the adaptation process, almost approaching the fully supervised upper bound.

%Aorta receives the highest Dice with 74.0\%. We think the underlying reason would attribute to its distinct geometric patterns and the clear boundaries on CT scans. In contrast, looking at the other three structures (i.e., LA-blood, LV-blood and LV-myo), the \textit{PaP-AdaNet} performances are not as high. The reason is that these anatomical structures are more challenging, given that they come with either relatively irregular geometrics or limited intensity contrast with the surrounding tissue. Nevertheless, by aligning the latent feature space of target domain to the source domain, our \textit{PaP-AdaNet} effectively bridges the performance gap by a very significant margin, at zero extra annotation cost.

\subsection{Ablation Studies on PnP-AdaNet Configurations}

\begin{table*}[t]
	\label{tab:abl-results}
	\centering
	%\small
	\fontsize{8pt}{2}
	\renewcommand{\arraystretch}{1.2}
	\caption{Domain adaptation Results of ablation studies adjusting configurations of the proposed PnP-AdaNet.}
	%\vspace{-4mm}
	\begin{center}
		\begin{tabular} { |m{2.2cm}|m{1.1cm}|m{1.1cm}|m{1.1cm}|m{1.1cm}|m{1.1cm}|m{1.1cm}|m{1.1cm}|m{1.1cm}|m{1.1cm}|m{1.1cm}|  }
			\hline
			Methods & \multicolumn{2}{c|}{AA} & \multicolumn{2}{c|}{LA-blood} & \multicolumn{2}{c|}{LV-blood} & \multicolumn{2}{c|}{LV-myo} &  \multicolumn{2}{c|}{Mean} \\
			\cline{2-11}
			&~~ Dice & ~~ ASD & ~~ Dice & ~~ ASD & ~~ Dice & ~~ ASD & ~~ Dice  & ~~ ASD &  ~~ Dice  & ~~ ASD \\
			\hline
			solely use $\mathcal{D}_f$                & 67.1$\pm$11.5 & 13.6$\pm$5.8 & 65.2$\pm$15.1 & ~9.3$\pm$4.3 & 58.1$\pm$8.5 & 15.9$\pm$7.6 & 48.0$\pm$4.7 & 15.0$\pm$5.1 & 59.6$\pm$10.0 & 13.5$\pm$5.7 \\
			$\mathcal{D}_m/\mathcal{D}_f$ rate = 0.1  & 74.0$\pm$7.3  & 12.8$\pm$3.2 & 68.9$\pm$5.2  & ~6.3$\pm$2.3 & 61.9$\pm$10.7& 17.4$\pm$7.0 & 50.8$\pm$7.0 & 14.7$\pm$4.8 & 63.9$\pm$7.5  & 12.8$\pm$4.3 \\
			$\mathcal{D}_m/\mathcal{D}_f$ rate = 0.2  & 71.1$\pm$8.8  & 14.2$\pm$5.5 & 67.5$\pm$17.3 & ~7.2$\pm$3.5 & 61.3$\pm$12.2& 18.0$\pm$7.5 & 47.3$\pm$9.5 & 17.6$\pm$5.8 & 61.8$\pm$11.9 & 14.2$\pm$5.6 \\
			$\mathcal{D}_m/\mathcal{D}_f$ rate = 0.3  & 72.0$\pm$10.7 & 14.8$\pm$4.2 & 65.7$\pm$16.0 & ~7.0$\pm$3.0 & 62.4$\pm$13.5& 15.3$\pm$7.2 & 49.2$\pm$6.1 & 13.4$\pm$4.7 & 62.3$\pm$11.6 & 12.6$\pm$4.8 \\
			$\mathcal{D}_m/\mathcal{D}_f$ rate = 0.4  & 73.3$\pm$6.2  & 12.5$\pm$6.0 & 68.2$\pm$21.4 & ~7.2$\pm$4.6 & 60.9$\pm$10.7& 18.2$\pm$7.4 & 52.6$\pm$10.1& 14.5$\pm$3.7 & 63.8$\pm$12.1 & 13.1$\pm$5.4 \\
			$\mathcal{D}_m/\mathcal{D}_f$ rate = 0.5  & 73.1$\pm$7.4  & 13.9$\pm$5.5 & 56.9$\pm$28.6 & ~8.3$\pm$3.5 & 50.0$\pm$13.4& 16.8$\pm$5.5 & 43.9$\pm$8.7 & 13.9$\pm$2.6 & 56.0$\pm$14.5 & 13.2$\pm$4.3 \\
			$\mathcal{D}_m/\mathcal{D}_f$ rate = 0.6  & 75.9$\pm$5.3  & 13.2$\pm$5.9 & 55.1$\pm$30.8 & ~7.6$\pm$4.8 & 48.7$\pm$5.0 & 20.2$\pm$7.3 & 43.7$\pm$7.3 & 17.9$\pm$6.0 & 55.9$\pm$12.1 & 14.7$\pm$6.0   \\
			$\mathcal{D}_m/\mathcal{D}_f$ rate = 0.7  & 63.2$\pm$6.4  & 17.5$\pm$5.9 & 62.2$\pm$15.2 & ~9.2$\pm$4.1 & 56.5$\pm$11.1& 17.3$\pm$6.5 & 44.2$\pm$2.9 & 17.2$\pm$4.9 & 56.5$\pm$8.9 & 15.3$\pm$5.3 \\
			\hline
			-- low-level layer       & 72.1$\pm$5.5   & 12.6$\pm$4.2  & 47.8$\pm$32.5 & ~9.9$\pm$6.3  & 50.5$\pm$12.6   & 16.0$\pm$4.9   & 48.7$\pm$8.6   & 15.0$\pm$3.3   &  54.8$\pm$14.8  & 13.4$\pm$4.7	 \\
			
			-- transaction layer     & 73.3$\pm$6.1	  & 12.9$\pm$4.0  & 58.2$\pm$17.7 & ~8.8$\pm$1.8  & 51.4$\pm$14.9   & 19.9$\pm$8.2   & 43.0$\pm$1.4   & 16.1$\pm$5.8   & 56.5$\pm$10.0   & 14.4$\pm$4.9   \\
			
			-- high-level layer      & 70.4$\pm$7.5   & 13.5$\pm$3.2  & 69.4$\pm$10.6 & ~6.6$\pm$1.8   & 53.6$\pm$14.2   & 21.8$\pm$8.0   & 43.0$\pm$5.0   & 18.2$\pm$6.9   & 59.1$\pm$9.3    & 15.0$\pm$5.0  \\
			\hline
		\end{tabular}
	\end{center}
	\vspace{-3mm}
\end{table*}

In this subsection, we extensively investigate the configurations of our \textit{PnP-AdaNet}.
% which are important components to make the framework works well.
Specifically, we observe the domain adaptation performance by adjusting two key properties: i) the ratio balancing these two discriminators $\mathcal{D}_m$ and $\mathcal{D}_f$, ii) the selected level of layers to input to the feature discriminator. We list the results of our experimented configurations in Table~2.

Specifically, we adjust the ratio of $\mathcal{D}_m/\mathcal{D}_f$ from $0.1\!:\!1$ to $0.7\!:\!1$, with a step of 0.1.
As we gradually increase this ratio, we are actually emphasizing more on the shape regularization from the mask discriminator. 
From Table~2, we first find that adding the mask discriminator as an auxiliary shape constrain is effective,
which helps to directly suppress those noises or shape distortions in the outputs.
Generally, the \textit{PnP-AdaNet} is not very sensitive to this ratio in a range of small-scale values.
When adjusting the radio from 0.1 to 0.4, the performance on segmenting CT images can stably achieve a Dice of over 60\%.
The change of the Dice score is not monotonic as we adjust the ratio.
Meanwhile, it also shows that the ratio could not be set too high, otherwise, the segmentation accuracy on the target domain would 
decrease as observed in our experiments.
The reason could be that straight-forward supervision of the mask is rather high-level.
The mask quality also heavily relies on the achieved domain-invariance from earlier features.
The two discriminators might also compete with each other when back-propagating their gradients to the generator, especially at the very beginning of training.
For the sake of a high adaptation performance, 
%aligning the latent feature space should be practically assigned as the major contributor in the learning process. 
monitoring the predicted masks in the compact semantic space should serve as an auxiliary supporting role.

As already observed that aligning the feature space is essential in our method, we furthermore investigate which layers to input their features to $\mathcal{D}_f$.
Intuitively, aligning shallow layers encourages domain-invariance of low-level features which are the basis for all higher-level representations.
The layer at the adaptation depth is obviously crucial as it is the dividing point in our plug-and-play setting.
Connecting the discriminator with deeper layers, although they are not updated in the adaptation process, can help to retain the alignment and prevent recovery of domain-specific information within the high-level space.
To observe the effectiveness of multi-level feature supervision, we conducted experiments by removing a low-level layer (RM4), the transaction layer (RM6) and a high-level layer (Conv10) from feature discriminator $\mathcal{D}_f$.
From Table~2, we can observe that missing either level of the feature maps would reduce the performance by a noticeable margin.
In practice, we also find that connecting the compact activations immediately before the softmax layer to $\mathcal{D}_f$ is crucial.
If we remove it from the feature discriminator input, a severe drop of performance will be seen.
This observation again confirms the importance of monitoring the high-level compact spaces with strong semantic relevance.

\subsection{Benchmarking at Cross-modality Cardiac Segmentation}

As our investigated topic of unsupervised domain adaptation on cross-modality medical image segmentation is quite new in the field,
there are few previous studies that could be compared with.
However, this problem deserves further careful explorations, given that CNNs are dominating current segmentation methods, and their generalization capability considerably matters.
Different from the cross-site and cross-sequence domain shifts, we regard cross-modality shift between CT and MRI as one of the most challenging settings.

\begin{figure}[t]
	\centering
	\includegraphics[width=0.49\textwidth]{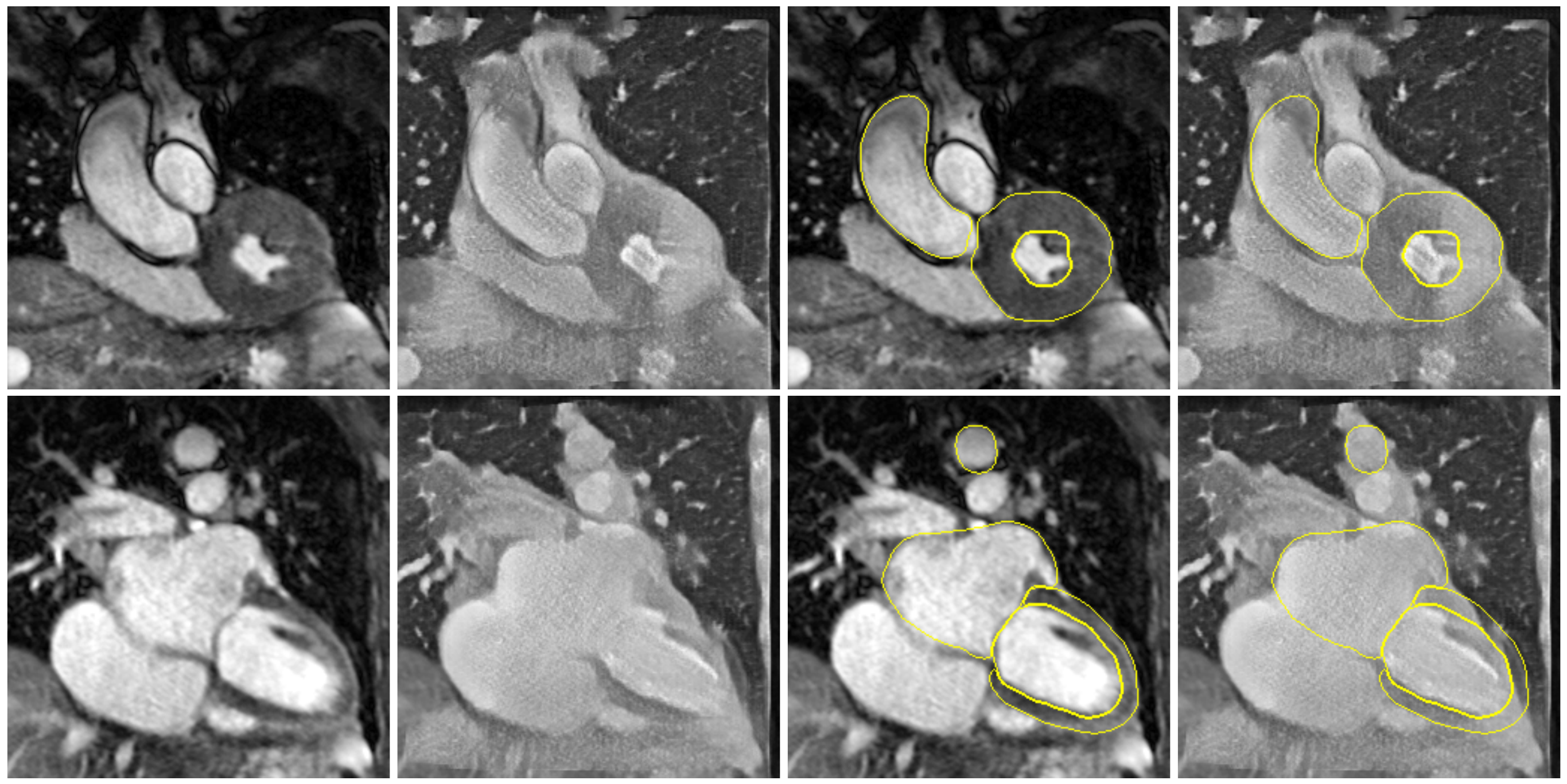}
	%\vspace{-6mm}
	\caption{Examples of MRI to CT image-to-image translations with CycleGAN. Left to right: original MRI image, generated CT image, MRI with segmentation ground truth, and generated CT with corresponding MRI ground truth.}
	\label{fig:cycleGAN}
	%\vspace{-2mm}
\end{figure}

To promote and facilitate future studies on cross-modality domain adaptation, in addition to our own prior work of~\cite{dou2018unsupervised}, we also implemented several state-of-the-art methods that are popular for unsupervised domain adaptation.
We used the same dataset and segmenter architecture settings for all the methods and introduce a benchmark as listed in Table~3.
Specifically, the approach of DANN~\cite{ganin2016domain} encourages domain-invariance in feature space.
% and the model contains a feature extractor, a label predictor and a domain classifier. 
The source and target domains share the feature extractor. A domain classifier is connected to the output of the encoder to monitor domain-invariance.
Another alternative approach is ADDA~\cite{tzeng2017adversarial} which also aligns source and target domain distributions in the feature space.
In this method, the source and target domains have their own feature encoders until the last softmax layer. 
A discriminator is used to differentiate which features come from which domain.
% This approach follows the GAN framework and uses a discriminator to differentiate which features come from which domain.
A third comparison method we include in the benchmark is CycleGAN~\cite{CycleGAN2017}, which produces impressive image-to-image transformations. We transform the MRI images to the appearance of CT, and train a segmenter with the transformed images and the MRI labels. We demonstrate typical examples of the generated CT images from MRI using CycleGAN in Fig.~\ref{fig:cycleGAN}.

\begin{table*}[t]
	\label{tab:benchmark}
	\centering
	\footnotesize
	\renewcommand{\arraystretch}{1.2}
	\caption{Benchmark of different methods on MRI to CT cross-modality domain adaptation of cardiac structure segmentation.}
	\begin{tabular} { |m{2.59cm}|m{1.1cm}|m{1.1cm}|m{1.1cm}|m{1.cm}|m{1.1cm}|m{1.1cm}|m{1.1cm}|m{1.1cm}| m{1.1cm}|m{1.0cm}|}
		\hline
		\multirow{2}{*}{Methods} & \multicolumn{2}{c|}{AA} & \multicolumn{2}{c|}{LA-blood} & \multicolumn{2}{c|}{LV-blood} & \multicolumn{2}{c|}{LV-myo} & \multicolumn{2}{c|}{Mean} \\
		\cline{2-11}
		& ~~~Dice & ~~ ASD & ~~~Dice & ~ ASD & ~~~Dice & ~ ASD & ~~~Dice  & ~ ASD & ~~~Dice  &  ~~ASD  \\
		\hline
		Seg-CT-noDA                        & 31.5$\pm$23.9 & 21.4$\pm$14.1 & ~2.7$\pm$0.8 & 19.6$\pm$4.6 &  ~3.4$\pm$5.8 & ~~~N/A      & 15.3$\pm$17.2 & 20.7$\pm$7.7  & 13.2$\pm$11.9 & ~~~N/A  \\
		\hline
		%DANN~\cite{ganin2016domain}        & 40.0$\pm$12.5 & 30.9$\pm$9.0 & 49.3$\pm$3.8  & 19.4$\pm$7.5 & 16.4$\pm$8.7 & 45.6$\pm$9.1  & 23.1$\pm$5.4 & 22.9$\pm$8.7 & 32.2$\pm$7.4 & 29.7$\pm$8.6 \\
		DANN~\cite{ganin2016domain}        & 39.0$\pm$35.1 & 16.2$\pm$5.8 & 45.1$\pm$23.6 & 9.2$\pm$2.9  & 28.3$\pm$11.8 & 12.1$\pm$1.6  & 25.7$\pm$13.2 & 10.1$\pm$3.2 & 34.5$\pm$20.9 & 11.9$\pm$3.4 \\
		ADDA~\cite{tzeng2017adversarial}   & 47.6$\pm$15.2 & 13.8$\pm$3.0 & 60.9$\pm$13.2 & 10.2$\pm$6.5 & 11.2$\pm$13.1 & ~~~N/A  & 29.2$\pm$16.4 & 13.4$\pm$5.0 & 37.2$\pm$14.5 & ~~~N/A\\
		CycleGAN~\cite{CycleGAN2017}       & 73.8$\pm$7.4  & \textbf{11.5$\pm$2.9} & \textbf{75.7$\pm$4.3} & 13.6$\pm$3.6 & 52.3$\pm$21.0 & \textbf{9.2$\pm$3.9} & 28.7$\pm$13.3 & ~\textbf{8.8$\pm$4.3} & 57.6$\pm$11.5 & \textbf{10.8$\pm$3.7}\\
		Prior PnP-AdaNet~\cite{dou2018unsupervised} & \textbf{74.8$\pm$6.2} & 27.5$\pm$7.6 & 51.1$\pm$11.2 & 20.1$\pm$4.5 & 57.2$\pm$12.4 & 29.5$\pm$11.7 & 47.8$\pm$5.8 & 31.2$\pm$10.1 & 57.7$\pm$8.9 & 27.1$\pm$8.5  \\ 
		\hline
		Proposed \textit{PnP-AdaNet}                & 74.0$\pm$7.3  & 12.8$\pm$3.2  & 68.9$\pm$5.2 & \textbf{6.3$\pm$2.3}  & \textbf{61.9$\pm$10.7} & 17.4$\pm$7.0 &\textbf{50.8$\pm$7.0}  & 14.7$\pm$4.8 & \textbf{63.9$\pm$7.5}  & 12.8$\pm$4.3 \\
		\hline
	\end{tabular}
	%\vspace{-3mm}
\end{table*}

Observing performances of the domain adaptation methods, we can see that all of them are able to effectively recover the segmentation accuracy on target CT data. 
%This means that, although the cross-modality domain adaptation is challenging, we can still address it by alignment either in the feature space or the image space. 
More specifically, in terms of different cardiac structures, all the methods perform better for ascending aorta and left atrium blood cavity than on the structures of left ventricle blood cavity and myocardium.
The reason could be that these anatomical structures, especially the myocardium, have relatively complicated geometry, which increases the difficulty for unsupervised domain adaptation. In contrast, the aorta presents a more compact shape as well as clear boundaries, and reasonably, receiving the best recovery.
Compared with DANN and ADDA which also used feature-level adaptation, our proposed \textit{PnP-AdaNet} achieves better performance.
The reason is that we neither share nor separate all the feature encoding layers between source and target domains.
Instead, our plug-and-play mechanism regards low-level features as domain-specific, while high-level feature encoders as sharable.
Moreover, connecting multi-level features to the discriminator also plays a crucial role.
The CycleGAN presents highly competitive results.
We analyze that this can attribute to the pixel-wise supervision, particularly towards the segmentation task.
The explicit supervision, although it maybe noisy, can encourage the predictions to consistently focus on the heart region.
The good ASD results with CycleGAN also indicate clean masks.
The deficiency of our feature-aligning method mainly appears at those unclear boundaries between neighboring structures, or wrong predictions on relatively homogeneous tissues but away from the ROI.
With a very simple post-processing strategy, i.e., only remaining the largest 3D connected component for every class, we can reduce the ASD to $4.1$, $5.4$, $7.4$ and $6.2$ for the AA, LA-blood, LV-blood and LV-myo, respectively.
Finally, compared with our own prior work of~\cite{dou2018unsupervised}, the performance improvement comes from the usage of dual discriminator, with more explicit constrain on the shape of segmentation masks.

Overall, all these methods, including ours, construct a strong benchmark on the task of cross-modality adaptation of cardiac segmentation on our dataset.

%Overall, we provide several baselines of unsupervised domain adaptation at the cross-modality cardiac segmentation task.
%Including ours, these methods make efforts from different perspectives, either in feature-level or image-level, and construct a strong benchmark on the delivered dataset.

\begin{table*}[t]
	\label{tab:results}
	\centering
	\footnotesize
	\renewcommand{\arraystretch}{1.2}
	\caption{Results of reverting the domain adaptation direction, i.e., adapting the segmentation network from CT to MRI.}
	%\vspace{-3mm}
	\begin{center}
		\begin{tabular} { |m{2.6cm}|m{1.1cm}|m{1.1cm}|m{1.1cm}|m{1.1cm}|m{1.1cm}|m{1.0cm}|m{1.1cm}|m{1.0cm}|m{1.1cm}|m{1.0cm}|  }
			\hline
			\multirow{2}{*}{Experimental settings} & \multicolumn{2}{c|}{AA} & \multicolumn{2}{c|}{LA-blood} & \multicolumn{2}{c|}{LV-blood} & \multicolumn{2}{c|}{LV-myo} &\multicolumn{2}{c|}{Mean}  \\
			\cline{2-11}
			&~~ Dice & ~ ASD & ~~ Dice & ~ ASD & ~~ Dice & ~ ASD & ~~ Dice  & ~ ASD & ~ Dice &  ~ ASD \\
			\hline
			Seg-MRI-noDA          & 31.5$\pm$23.9 & 21.4$\pm$14.1& ~2.7$\pm$0.8 & 19.6$\pm$4.6 & ~3.4$\pm$5.8  & ~~~N/A      & 15.3$\pm$17.2 &  20.7$\pm$7.7 & 13.2$\pm$11.9 & ~~~N/A\\
			
			\hline
			Proposed \textit{PnP-AdaNet}   & 43.7$\pm$10.8 & 11.4$\pm$3.2 & 47.0$\pm$7.3 & 14.5$\pm$4.1 & 77.7$\pm$10.4 & 4.5$\pm$1.4 & 48.6$\pm$2.9  & ~~5.3$\pm$1.8 & 54.3$\pm$7.9 & 8.9$\pm$2.6 \\
			
			\hline
		\end{tabular}
	\end{center}
	\vspace{-4mm}
\end{table*}

\subsection{Reverting Domain Adaptation Direction}

A natural question to ask is whether domain adaptation for the segmentation task is \textit{symmetric to modality}, \textit{i.e. }whether the reverse adaptation direction from CT to MRI can also be achieved.
To investigate this, we apply the same model setting, but only replacing the source domain as CT and target domain as MRI. 
The quantitative adaptation results are presented in Table~4. Directly using the CT segmenter on MRI data also unsurprisingly fails.
Our proposed \textit{PnP-AdaNet} is able to recover the average segmentation Dice to 54.3\%.
Notably, the best recovered structure in this reverse setting is the LV-blood, not the same structure (AA) as the direction of MRI to CT. 
As a more interesting observation, the Dice increases from complete failure (3.4\%) to a considerably high value (77.7\%).
Results of the other three structures are not as promising.
Compared with adaptation from MRI to CT, the reverse direction yields lower performance, which can be expected.
%The image quality (such as noise level and boundary sharpness) of CT data is generally better than that of MRI.
Segmenting cardiac MRI itself is more difficult than segmenting cardiac CT.
This is also notable from Table~1, where the CT segmentation Dice is higher than the result of MRI across all four structures.
In these regards, transferring a CT segmenter to MRI seems more challenging.
Our experimental results indicate that cross-modality domain adaptation can be achieved in both directions, however, the difficulty seems asymmetric.

\section{Discussions and Conclusions}
\label{sec:dis}

%\subsection{Overview of significance of the paper}

In this paper, we try to tackle the domain adaptation problem under a very challenging as well as important setting of cross-modality medical datasets.
%As deep learning networks have nowadays been the fundamental methodologies for solving the segmentation and detection tasks in the field, 
As deep learning has become the de-facto standard for solving segmentation and detection tasks, investigating its generalization capability and robustness is essential.
Existing successful practice using deep networks is to train and test the models with the same data source.
However, it has been frequently revealed in very recent works, that the models would perform poorly on unseen datasets~\cite{karani2018lifelong,gibson2018inter,chen2018semantic}.
Resolving the domain adaptation issue holds great potentials for, applying trained deep learning models to wider clinical use, building more powerful networks using large-scale database combing images from multiple sites, and helping to understand how the networks capture the data distributions to make recognition predictions.
In these regards, we explore effective unsupervised domain adaptation solution, under one of the most challenging settings as the \textit{multi-class segmentation of cross-modality medical images}.

%\subsection{difficulty of cross-modality domain adaptation}
The appearance differences between MRI and CT scans are apparent.
Although the human eye could match the structures on the two modalities, from a perspective of image computing, the system ``sees" distinct distributions of intensity values.
For concrete example, see Fig.~1 again, the intensity range of myocardium and its contrast with the nearby tissues is very different on MRI and CT.
In this case, the model that learns discriminative features from one modality is naturally inapplicable to the data manifold of another modality.
Generally speaking, it is rather difficult for a model to generalize from one domain to another, where the measure of their support intersection is almost zero.
%Although it is highly challenging, we still have to conquer the difficulties, as 
On the other hand, multi-modality data have become indispensable tools in modern clinical routine. Cross-modality synthesis and segmentation has been gaining research popularity rapidly~\cite{JoRobust,hiasa2018cross}.
Considering cardiology as currently the most typical scenario which uses both MRI and CT, we conduct our study on cardiac images.

%\textcolor{blue}{(xiahai, pls notice above para, do we need to strengthen clinical significance for usage of cross-modality cardiac scan?)}

%\subsection{Our proposed method highlight and adaptation depth}

Towards domain adaptation, we present a novel framework, i.e., \textit{PnP-AdaNet}, which is very flexible and light-weight in practical use.
We replace the early layers of an established network with a DAM at the testing process.
The assumption is that, in cross-modality medical images for the same organ, those low-level features are domain-specific, while the semantic feature compositions at higher layers could be re-used across different modalities.
To learn the DAM, we encourage the extracted features of the target domain to be aligned with the source feature distributions, throughout the network. A notable setting in our \textit{PnP-AdaNet} is incorporating multi-level adversarial learning, which makes the DAM tightly conditioned by the source distributions.
Another hyper-parameter, which needs to be considered when using the plug-and-play strategy, is the adaptation depth $d$.
In our prior work of~Dou et al.~\cite{dou2018unsupervised}, we have experimentally investigated the influence of adjusting the choice of adaptation depth.
Intuitively, a smaller $d$ (i.e., shallower DAM with fewer parameters) might be less capable of learning effective feature mappings across domains.
On the other hand, training a much deeper DAM solely with adversarial gradients would be difficult and unstable.
A recent work studying a generalization upper-bound for unsupervised domain adaptation also indicated similar insights~\cite{galanti2018generalization}.
The paper derived that minimizing the generalization risk across domains requires a function with particular characteristics.
It needs to be capable enough for reducing the Wasserstein distance to a desired degree for maintaining adaptation quality, meanwhile, also should control the complexity of reducing the generalization risk.
Our experimental findings in~\cite{dou2018unsupervised} matched these insights, i.e., plugging the first six residual blocks yielded the highest performance. Hence, we suggest to select a middle-level layer as the adaptation depth, when embedding our plug-and-play strategy into other similar applications.

%\subsection{feature-level adaptation v.s. image-to-image translation}

When thinking of solutions for domain adaptation, an alternative way is to take advantages of image-to-image translation.
The networks using cycle-consistency loss can generate plausible synthetic images, which then can either be employed for training or testing.
The potential risk of pixel-space translation would be that the distortions and artifacts might be propagated or amplified in downstream processing.
For example, Cohen et al.~\cite{cohen2018distribution} found that CycleGAN based medical image translations trained on imbalanced datasets would hide dangerous brain tumors in synthetic images.
From our own experimental experiences, the cross-modality translation of non-tumor medical images is feasible using CycleGAN, and the reconstructed images also look quite realistic. It can bring positive effects when using synthetic images towards the scenario of domain adaptation, however, it cannot solve the problem perfectly. We analyze that the reasons are at least two-folds. One is that there still exists domain shift between the synthetic image and real image, although their appearances somehow look similar~\cite{zhang2018translating}. 
The other is that the generated image may on a pixel-level not match with the original image, therefore, the training pairs of image and label are imperfect.
In contrast, feature-level adaptations, such as ours, do not suffer from these problems, as the adaptation and task-specific training are not detached.
%The existing researches reveal that those feature-level and image-level adaptations tackle the task from different perspectives. We would say that they are in fact complementary to each other, as the feature alignment builds domain-invariant latent space, while the image translation can explicitly provide pixel-to-pixel supervision for segmentation task. For the future work, we recognize combing feature-level adaptations with image-to-image translations in an end-to-end learning model as a highly promising direction.

%\subsection{2D vs 3D, limitation of the work}
Although we achieved promising results, there is still limitation.
The backbone of our network is 2D CNN. In practice, we aggregate the adjacent three slices as the input channels to the models.
%This is common practice when the community was starting to use CNN for processing volumetric medical data.
This is common practice to overcome current single GPU memory constraints.
%Currently, people usually 
It maybe beneficial to employ more carefully tailored networks for CT/MRI image analysis with 2.5D~\cite{roth2014new} or 3D CNNs~\cite{dou2016automatic}.
In this work, we also tried to conduct the domain adaptation on the basis of 3D segmentation networks. However, we were faced with difficulties of memory consumption given the use of one segmenter, one generator and two discriminators.
Moreover, optimizing 3D networks in unsupervised adversarial learning is also very challenging.
In these regards, we chose to first down-grade the network to 2D, so that we can focus on the domain adaptation part which is the core of this paper.
%We set our basis as the 2D dilated network which reached standard performance compared with the state-of-the-arts on the cardiac dataset.
%In up-to-date literatures, 
There are still not many works using 3D CNNs for GANs on medical applications.
One recent work is Zhang et al.~\cite{zhang2018translating} which synthesized CT from MRI on cardiac data using an end-to-end 3D CNN.
%Another work is Jin et al.~\cite{jin2018ct} which used a 3D conditional GAN to simulate lung nodules in CT. 
Another work is Pan et al.~\cite{pan2018synthesizing} which used a 3D cycle-consistent GAN to synthesize missing PET from MRI for neuroimage.
To the best of our knowledge, the work of~\cite{kamnitsas2017unsupervised} was the first to employ 3D CNNs in domain adaptation for medical image segmentation task, which tackled different MRI sequences.
Exploring the effectiveness of cross-modality domain adaptation approaches on the basis of 3D network is planned in our future work.

%\subsection{Conclusion}

In conclusion, we investigate the challenging yet crucial task of unsupervised domain adaptation on cross-modality medical images.
We propose a novel approach, called \textit{plug-and-play adversarial domain adaptation network (PnP-AdaNet)}, which aligns the latent feature space of the target domain to that of the source domain. Our extensive experiments have validated the effectiveness of our model.
Moreover, to facilitate future research on the cross-modality domain adaptation problem, we contribute a benchmark on the specific application of MRI/CT cardiac structure segmentation with a well-organized database.
We believe that the cross-modality domain adaptation task will witness rapid developments.

%\section{Discussion Ouyang}
%\input{discussion_ouyang_v2.tex}

%\newpage

\bibliographystyle{IEEEtran}
\bibliography{refs}

\end{document}